\documentclass[letterpaper]{article} 
\usepackage{aaai25}  
\usepackage{times}  
\usepackage{helvet}  
\usepackage{courier}  
\usepackage[hyphens]{url}  
\usepackage{graphicx} 
\usepackage{natbib}  
\usepackage{caption} 
\usepackage{algorithm}
\usepackage{algorithmic}

\usepackage{subcaption}
\usepackage{multirow}
\usepackage{amsmath}
\usepackage{mathrsfs}
\usepackage{amssymb}
\usepackage{amsfonts}
\usepackage{url}            
\usepackage{booktabs}
\usepackage{pifont}
\usepackage{tikz}

\newcommand*\circled[1]{\tikz[baseline=(char.base)]{
            \node[shape=circle,draw,inner sep=0.4pt] (char) {#1};}}

\usepackage{newfloat}
\usepackage{listings}
\DeclareCaptionStyle{ruled}{labelfont=normalfont,labelsep=colon,strut=off} 
\floatstyle{ruled}
\newfloat{listing}{tb}{lst}{}
\floatname{listing}{Listing}
\title{Unlocking the Power of Patch: Patch-Based MLP for Long-Term Time Series Forecasting}
\author {
    Peiwang Tang\textsuperscript{\rm 1, \rm 2},
    Weitai zhang\textsuperscript{\rm 1, \rm 2}
}
\affiliations {
    \textsuperscript{\rm 1}iFLYTEK Research, Hefei, China \\
    \textsuperscript{\rm 2}University of Science and Technology of China, Hefei, China
    \\
    tpw@mail.ustc.edu.cn, zwt2021@mail.ustc.edu.cn
}

\usepackage{bibentry}

\begin{document}

\maketitle

\begin{abstract}
Recent studies have attempted to refine the Transformer architecture to demonstrate its effectiveness in Long-Term Time Series Forecasting (LTSF) tasks. Despite surpassing many linear forecasting models with ever-improving performance, we remain skeptical of Transformers as a solution for LTSF. We attribute the effectiveness of these models largely to the adopted Patch mechanism, which enhances sequence locality to an extent yet fails to fully address the loss of temporal information inherent to the permutation-invariant self-attention mechanism. Further investigation suggests that simple linear layers augmented with the Patch mechanism may outperform complex Transformer-based LTSF models. Moreover, diverging from models that use channel independence, our research underscores the importance of cross-variable interactions in enhancing the performance of multivariate time series forecasting. The interaction information between variables is highly valuable but has been misapplied in past studies, leading to suboptimal cross-variable models. Based on these insights, we propose a novel and simple Patch-based MLP (PatchMLP) for LTSF tasks. Specifically, we employ simple moving averages to extract smooth components and noise-containing residuals from time series data, engaging in semantic information interchange through channel mixing and specializing in random noise with channel independence processing. The PatchMLP model consistently achieves state-of-the-art results on several real-world datasets. We hope this surprising finding will spur new research directions in the LTSF field and pave the way for more efficient and concise solutions. Code is available at: \url{https://github.com/TangPeiwang/PatchMLP}
\end{abstract}

%

\section{Introduction}
\begin{figure*}[ht]
	\centering
	\includegraphics[scale=0.36]{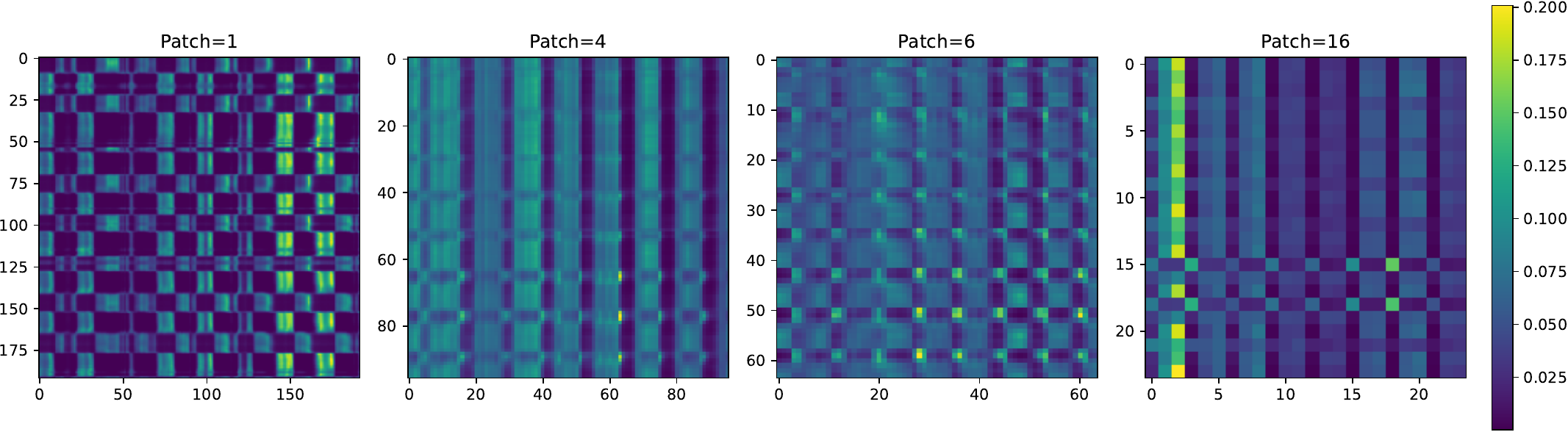}
  \caption{The self-attention scores of a 2-layer Transformer with different Patch size trained on ETTh1. We follow the setup of PathcTST \cite{nie2022time}, retaining only the Encoder while replacing the Decoder with a simple MLP, and using a channel independent approach. A patch size of 1 is equivalent to the original Transformer, indicating that time series data often exhibits a trend of being segmented into patches \cite{zhang2022crossformer,tang2023infomaxformer}, and an increase in patch size can mitigate this to some extent.}
  \label{atention_1}
\end{figure*}

Long-term Time Series Forecasting (LTSF) is a critical area of research in the fields of statistics and machine learning, aimed at using historical data to predict the future of one or more variables over a certain period \cite{oreshkin2019n,gong2023patchmixer,lin2023petformer}. Time series data are organized in chronological order and reveal underlying dynamic patterns, be they cyclical or non-cyclical \cite{cryer1986time}. Such forecasts play a vital role in various sectors, including biomedicine \cite{liu2018learning}, economics and finance \cite{patton2013copula}, electricity \cite{zhou2021informer}, and transportation \cite{yin2016multivariate}.

Multivariate Time Series (MTS) consist of multiple variables recorded at the same time point, where each dimension may represent an individual univariate time series or be considered as a signal with multiple channels. With the advancement of deep learning, a number of models have been developed to enhance the performance of MTS forecasting \cite{borovykh2017conditional,bai2018empirical,liu2020dstp}. In particular, recent models based on the Transformer \cite{vaswani2017attention} architecture have demonstrated significant potential in capturing long-term dependencies \cite{radford2018improving,devlin2018bert}. In recent years, Transformers have emerged as a leading architecture in time series forecasting \cite{lim2021temporal,liu2022non,shen2023take}, initially applied in the field of Natural Language Processing (NLP) \cite{radford2019language,brown2020language} and subsequently expanded to other domains such as Computer Vision (CV) \cite{liu2021swin,feichtenhofer2022masked} through the adaptation of the \textbf{Patch} method, becoming a versatile framework.

Originally, these models utilized a \textbf{channel mixing} approach \cite{li2023mts}, which involves projecting vectors from different channels recorded at the same time point into an embedding space and blending the information \cite{zhou2021informer,tang2023infomaxformer} ,and many models adopt the idea of time series decomposition \cite{wu2021autoformer,zhou2022fedformer,wang2023timemixer}. 
Nonetheless, recent studies have shown that a \textbf{channel independence} model might be more effective \cite{nie2022time,han2023capacity}. 
Channel independence means that each input token contains information from only one channel, and intuitively, the models treat the MTS as separate univariate series for individual processing. 
These studies suggest that for LTSF tasks, an emphasis on channel independence may be more effective than channel mixing strategies \cite{gong2023patchmixer}.

Moreover, the effectiveness of Transformers in LTSF tasks has been challenged by a reconsideration of the Multi-Layer Perceptron (MLP) \cite{zeng2023transformers,ekambaram2023tsmixer,chen2023tsmixer}, whose surprisingly simple architecture has outperformed all previous Transformer models. This raises a compelling question: Are Transformers effective for LTSF tasks? In response to this skepticism, recent Transformer-based models have adopted patch-based representations and achieved noteworthy performance in LTSF \cite{nie2022time,chen2024pathformer}.

This study raises three critical inquiries: 
\begin{itemize}
 \item Is the method of channel mixing truly ineffective in MTS forecasting?
  \item Can simply decomposing the original time series truly better predict the trend and seasonal components?
 \item Does the impressive performance of the Patch-based Transformer derive from the inherent strength of the Transformer architecture, or is it merely due to the use of patches as the input representation?
\end{itemize}

\begin{figure*}[htbp]
\centering
\begin{subfigure}{0.3\linewidth}
  \centering
  \includegraphics[scale=0.28]{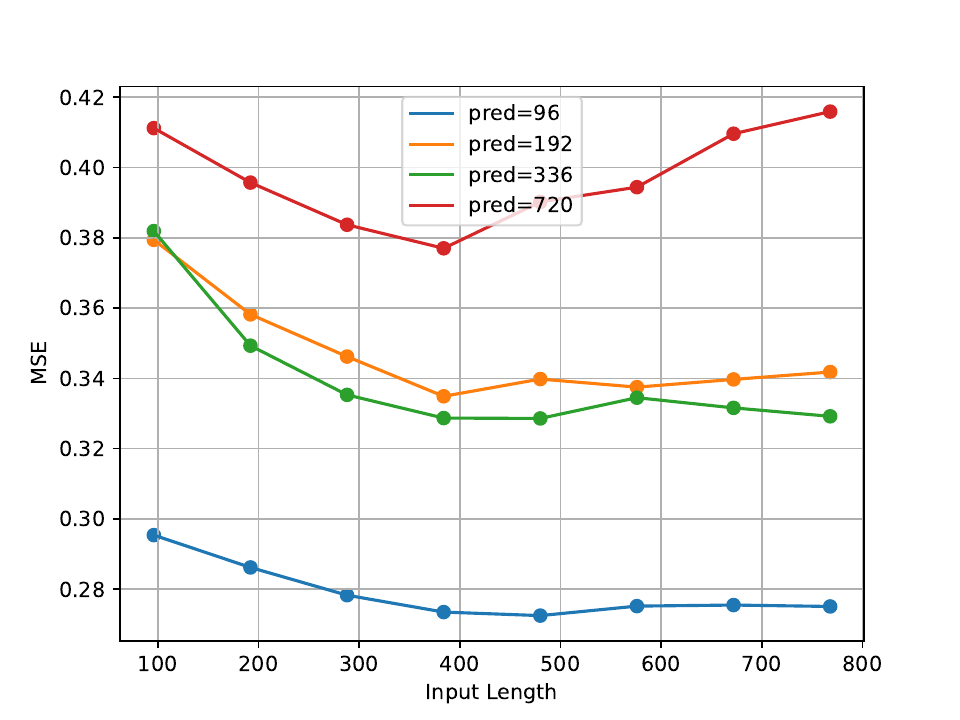}
      \caption{}
  \label{fig11}
\end{subfigure}%
\begin{subfigure}{0.32\linewidth}
  \centering
  \includegraphics[scale=0.28]{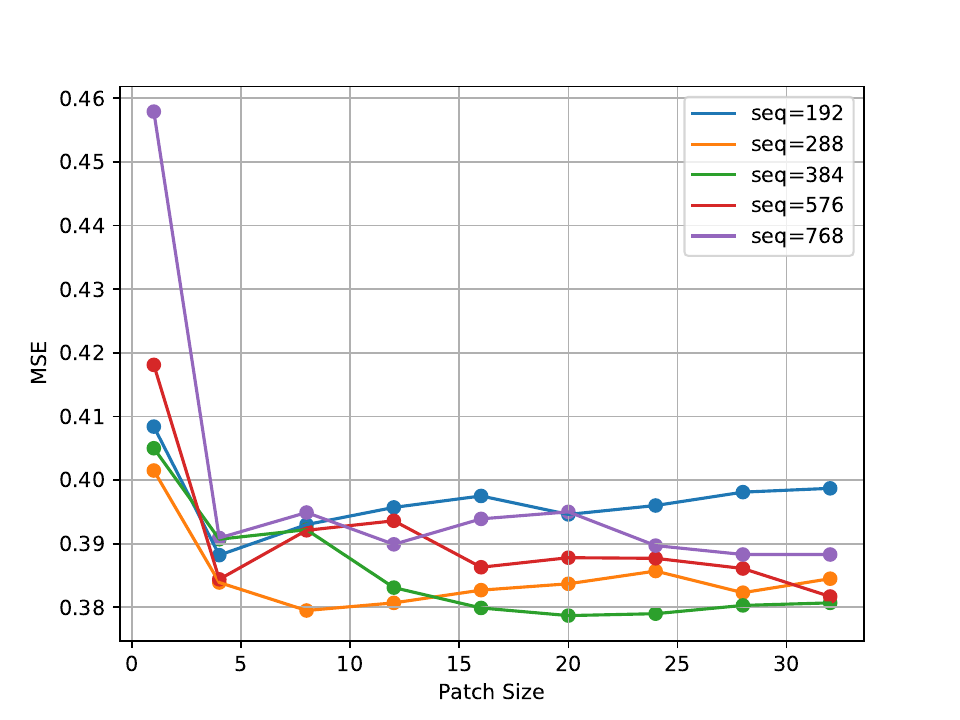}
      \caption{}
  \label{fig12}
\end{subfigure}%
\begin{subfigure}{0.32\linewidth}
  \centering
  \includegraphics[scale=0.28]{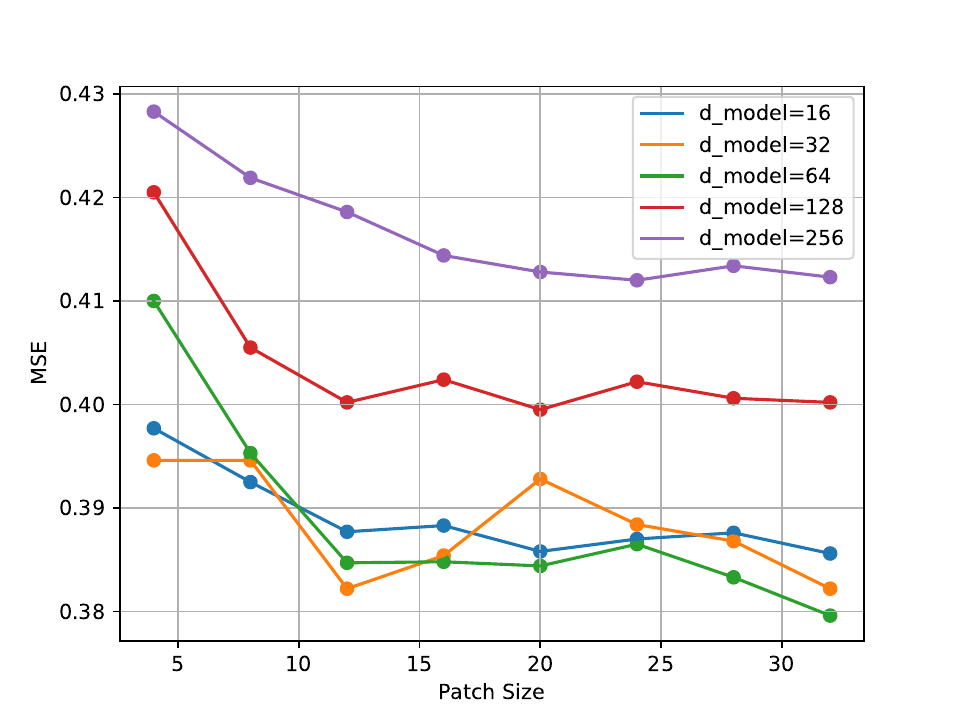}
      \caption{}
  \label{fig13}
\end{subfigure}
\caption{Experimental results of Patch Transformer on the ETTh2 dataset. 
(a) Maintain all other parameters constant, and present the MSE outcomes for four forecast lengths with only the input length altered.
(b) Keeping all other parameters constant and only altering the size of the patch, the MSE results for a forecast length of 720 with five different input lengths.
(c) Maintaining all other parameters unchanged and only varying the patch size, the MSE results for five different $d_{model}$ values with both input and forecast lengths set to 720.}
\label{fig1}
\end{figure*}

In this paper, we specifically discuss the three issues mentioned above and introduce the PatchMLP, a novel and concise Patch-based MLP model tailored for LTSF task.
PatchMLP is entirely based on fully connected networks and incorporates the concept of patch. Specifically, we employ a patch methodology to embed the original sequences into representational spaces, then extract the smooth components and noisy residuals of the series using a simple moving average technique for separate processing. We independently process stochastic noise across channels and facilitate semantic information interchange between variables through channel mixing. Upon evaluation on numerous real-world datasets, the PatchMLP model demonstrated state-of-the-art (SOTA) performance. To be precise, our contributions can be summarized in three aspects:
\begin{itemize}
 \item We analyze the effectiveness of Patches in time series forecasting and propose a Multi-Scale Patch Embedding (MPE) approach. Unlike previous embedding methods that used a single linear layer, MPE is capable of capturing multiscale relationships between input sequences.
 \item We introduce a novel entirely MLP-based model, named PatchMLP. By utilizing moving averages, it performs the decomposition of latent vector and adopts a different approach to channel mixing for semantic information exchange across variables.
 \item Our experiments across a wide range of datasets in various fields show that PatchMLP consistently achieves SOTA performance across multiple forecasting benchmarks. Moreover, we conduct an extensive analysis of Patch-based methods with the aim of charting new directions for future research in time series forecasting.
\end{itemize}

\section{Patch for Long-Term Time Series Forecasting}

\textbf{Why is Patch effective in time series forecasting}?  \cite{lee2023learning, zhong2023multi} 
We have studied a patch-based Transformer model adopting channel independent mechanisms, and we set the decoder layer as a simple single-layer MLP for time series forecastine \cite{nie2022time}.
As shown in the Figure ~\ref{fig12}, we can clearly see that for the same input length, as the size of the patch increases, the overall Mean Squared Error (MSE) of the model exhibits a trend of decreasing and then increasing, or decreasing and then stabilizing. As the input length increases, the patch size required for the model to achieve optimal performance also gradually increases. In the original input data, the effect of Attention is so poor that it cannot even outperform a simple single-layer MLP, which raises the question of whether Attention might not be the optimal choice for time series modeling to some extent, as it cannot handle long-term time series well.

Compared to textual data in the NLP field, original time series data contains many redundant features due to high-frequency sampling, and is easily influenced by noise to some extent during the sampling process \cite{tang2022mtsmae}. We believe that for data with excessive noise in the original data, the original Transformer's Attention mechanism is not the optimal choice as it cannot effectively eliminate the noise. Sparse Attention performs better because this sparse mechanism can effectively mitigate the impact of noise to some extent \cite{li2019enhancing,wu2020adversarial,liu2021pyraformer}, but it may also attenuate the impact of original features.

Patch, on the other hand, compresses the data, reduces the dimensionality of the input data, and decreases redundant features. Additionally, patch provides a certain degree of smoothing, which can reduce the influence of outliers to some extent, and help filter out fluctuations and random noise, retaining more stable and representative information. In the CV field, patch also works well in ViT \cite{dosovitskiy2020image}, MAE \cite{he2022masked}. Time series data usually contains patterns at different scales, and patch provides a modeling of short-term time series, enhancing local information in the sequence, allowing the model to better learn and capture local features. Therefore, we believe that the effectiveness of Transformer is not due to the effect of Attention, but rather due to the presence of patch.

\begin{figure*}[ht]
	\centering
	\includegraphics[scale=0.7]{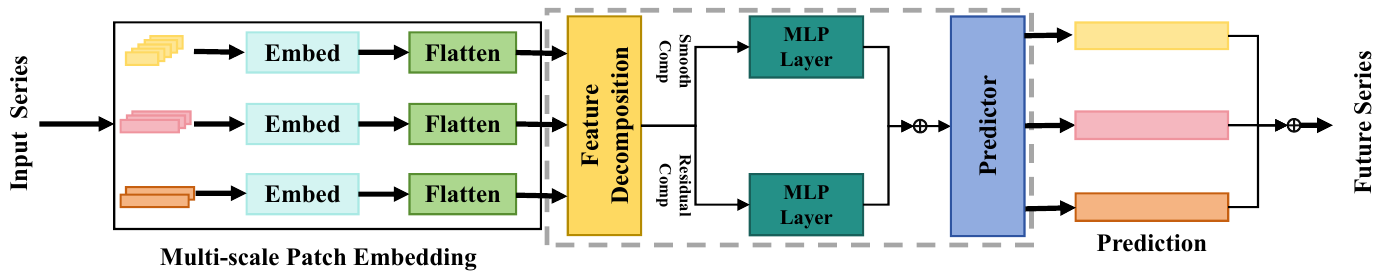}
  \caption{Overall structure of PatchMLP. First, the raw time series of different variables are independently processed through Multi-scale Patch Embedding. Then, Feature Decomposition uses moving averages to decompose the embedded tokens into smooth components and noisy residues. Next, a MLP processes the sequences in two ways: intra-variable and inter-variable. Finally, the Predictor maps the latent vectors back to predictions and aggregates them into future series.}
\label{PatchMLP}
\end{figure*}

So, is a larger patch always better? In Figure ~\ref{fig12}, a larger patch does not seem to achieve better results, which may be partly influenced by the size of the hidden layer ($d_{model}$). With a larger patch size, more time points are projected into a fixed-size $d_{model}$, which may lead to excessive compression. As shown in Figure ~\ref{fig13}, as $d_{model}$ increases, the performance of the model gradually improves, indicating that larger patches perform better. In the extreme case, a patch that treats the entire input sequence as one patch and projects it into a vector is similar to iTransformer \cite{liu2023itransformer}. However, a larger $d_{model}$ indicates that there are more parameters for the model to learn, which can easily lead to underfitting and result in decreased model performance.

\section{PatchMLP}

The problem of multivariate time series forecasting is to input the historical observations $\mathcal{X}=\left \{x_{1}, \cdots ,x_{L} |x_{i}\in \mathbb{R}^{M} \right \}$, and the output is to predict corresponding future sequence $\mathcal{X}=\left \{x_{L + 1}, \cdots ,x_{L+T } |x_{i}\in \mathbb{R}^{M} \right \}$, where $L$ and $T$ are the lengths of input and output sequences respectively, and $M$ is the dimension of variates.

As shown in Figure ~\ref{PatchMLP}, PatchMLP consists of four network components:  Multi-Scale Patch Embedding layer, Feature Decomposition layer, Multi-Layer Perceptron (MLP) layer, Projection layer. The Multi-Scale Patch Embedding layer embeds the multivariate time series into latent space. The Feature Decomposition layer decomposes the latent vector into the smooth components and noisy residuals, then operate separately through MLP layer. Finally, the latent vector are mapped back to the feature space through the Projection layer to obtain the future sequences $\hat{\mathcal{X}}$. Next, we will introduce the above modules separately.

\subsection{Multi-Scale Patch Embedding}
Time series analysis relies on the accurate identification of local information within the sequence and optimization of the model. The use of patches can provide the model with a local view of the time series in the short term, thereby enhancing the representation of local information within the sequence and enabling the model to learn and capture these local features more precisely. However, traditional methods typically employ patches of a single-scale to embed the raw time series, which makes the model inclined to learn single-scale temporal relationships while neglecting the multi-scale nature and complexity of time series data.

In practical applications, the single-scale patch strategy may lead to the model capturing inaccurate or incomplete local features. This is because the data characteristics of different time series often have variation, and a patch of the same scale cannot universally adapt to all types of time series. For example, a time series containing multiple cyclical patterns, a single-scale patch cannot effectively identify and learn the cyclical features present at different frequencies.

To capture local information in more detail and to fully understand the temporal relationships within the time series, we adopt multi-scale patch. This includes shorter patch to capture local high-frequency patterns, as well as longer patch to unearth long-term seasonal, trends and periodic fluctuations. Through the introduction of multi-scale patch, the model can flexibly learn representative features over different lengths of time spans, thus enhancing predictive accuracy and model generalization capabilities.

To decompose the multivariate time series $\mathcal{X}$ into a univariate series $x$, we have developed a suite of patches $\mathcal{P}$ across various scales to process $x$. 
For a particular scale $p \in \mathcal{P}$, $x$ is first divided into non-overlapping patches where the length of patch corresponds to $p$, thus the patching process yields the sequence of patches $ x_p \in \mathbb{R}^{N \times p}$, where $N$ is the number of patches. 
Subsequently, we employ a single-layer linear layer to embed the patches $x_p$, resulting in latent vectors $x_e \in \mathbb{R}^{N \times d}$, $d$ is the embedding dimension, the $d$ corresponding to different scales patches can be different. 
These latent vectors are then unfolded to obtain the final embedding vectors $X \in \mathbb{R}^{1 \times d_{model}}$, $d_{model}$ is the final embedding dimension input into the model.
This multi-scale patch strategy permits the model to capture and learn the intrinsic dynamics of time series at different levels, thereby providing more precise insights when forecasting future trends and patterns.
\subsection{Feature Decomposition}
Time series often contain complex temporal patterns, including seasonal fluctuations, trends, and other irregular influencing factors. Many studies attempt to decompose time series into these fundamental components through decomposition methods, to achieve better understanding and forecasting. Although decomposition is a powerful tool, traditional methods may encounter some difficulties when dealing with the raw sequences, especially when there are complex and mixed patterns in the sequences. These methods often struggle to accurately separate clear trends and seasonal components, particularly when the data contains substantial noise or nonlinear components.

In response to this issue, we propose a different idea, instead of directly decomposing the original sequence, we decompose the sequence's embedding vector. Embedding vector are representations formed by mapping time series to a high-dimensional space, and these representations often capture the core information of the original data. By decomposing embedding vector, we can distinguish smoother components and noise-containing residuals, which helps eliminate the interference of random fluctuations on analysis and forecasting. In practice, we use the operation of Average Pooling ($AvgPool$) to smooth the time series, aiming to reduce random fluctuations and noise in the data. Moreover, in order to maintain the length of the time series unchanged while smoothing, we applied a padding operation.
\begin{equation}
	\begin{split}
		X_s & = AvgPool(X) \\ 
		X_r &= X - X_s
	\end{split}
\end{equation}
$X_s$ and $X_r$ respectively represent the extracted smooth components and residual components. By operating on embedding vector rather than the original sequence, the model can extract and identify the fundamental components of time series more precisely, enabling a better understanding of the intrinsic structure of time series, and ultimately, improving the accuracy of time series forecasting.
\begin{figure*}[ht]
	\centering
	\includegraphics[scale=0.78]{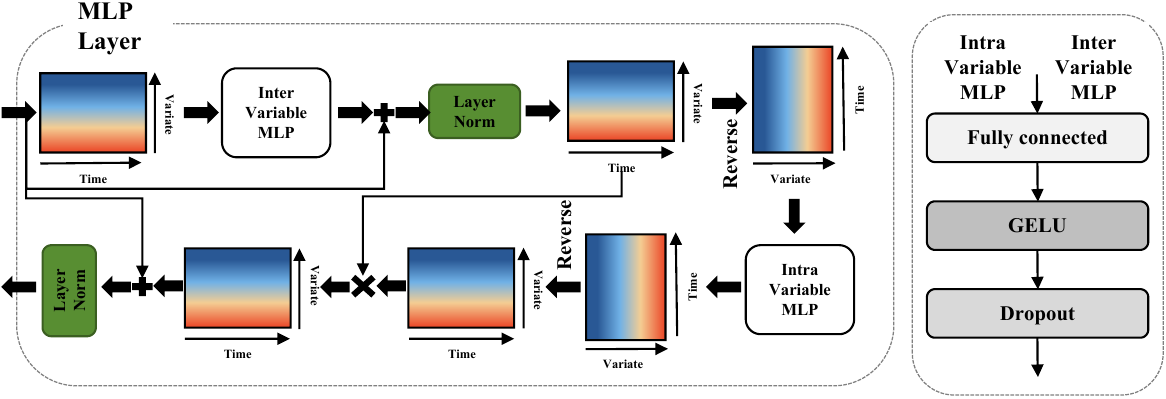}
  \caption{Overall structure of MLP layer. The embedded vectors first interact with the temporal information within the variable through the Intra-Variable MLP. Then interact with the feature domain information between variables through the Intra-Variable MLP. Subsequently, they are multiplied by the input of the Inter-Variable MLP using a dot-product approach. Finally, they are added to the initial input of the MLP Layer using skip connections.}
\label{MLP}
\end{figure*}

\subsection{MLP layer}
In the context of MTS forecasting, the MLP layer alternately applies MLPs within Intra-Variable (time domain) and Inter-Variable(feature domain) to enhance predictive performance. The specific architecture, as depicted in Figure ~\ref{MLP}, can be described in detail as follows:

\textbf{Intra-Variable MLP:} This component is focused on the identification of time-correlated patterns within time series. Specifically, it utilizes a network architecture composed of fully connected layers, nonlinear activation functions, and dropout. Parameters are applied in the temporal domain and are shared across inter-variable. Simple linear models are easy to understand and implement, and have been proven to learn complex time patterns and capture more complex time dependencies.

\textbf{Inter-Variable MLP:} 
Complementary to the intra-variable MLP,
the inter-variable MLP aims to model the mutual influences among MTS variables.
This component also consists of fully connected layers, activation functions, and dropout, applying fully connected layers in the feature domain to achieve parameter sharing within the intra-variable.
To enhance the cross-variable interactivity, we utilize a dot product mechanism, integrating the dot product results between the MLP outputs and inputs, which enhances the model's nonlinear representational capability.

\textbf{Residual Connections:} Residual connections \cite{he2016deep} are applied after each MLP, enabling the model to learn deeper architectures more efficiently. These connections not only help to mitigate the problems of vanishing gradients in deep networks but also provide the model with a shortcut path that ensures key temporal or feature information is not overlooked.


While the minimalist architecture of PatchMLP may not appear as eye-catching as some of the recently proposed, more complex Transformer models, empirical results indicate that PatchMLP exhibits competitive performance, both in terms of training efficiency and accuracy on standard benchmark tests, relative to SOTA models.

\subsection{Loss Function}

Our loss function is calculated by the Mean Square Error (MSE) between the model prediction $\hat{x}$ and the ground truth $x$:
$
\mathcal{L} = \frac{1}{M}  \sum_{M}^{i=1} \left \| \hat{x}^i_{L+1:L+T} - x^i_{L+1:L+T} \right \|^2_2 
$
and the loss is propagated back from the Projection output across the entire model.

\section{Experiments}
We comprehensively evaluated the proposed PatchMLP in various time series forecasting applications, verifying the generality of the proposed framework, and have further delved into the investigation of the effectiveness of individual components of PatchMLP when applied in LSTF tasks.

\textbf{Datasets:}  
We evaluated the performance of our proposed PatchMLP on 8 commonly used LSTF benchmark datasets: Solar Energy \cite{lai2018modeling}, Weather, Traffic, Electricity (ECL) \cite{wu2021autoformer}, and 4 ETT datasets (ETTh1, ETTh2, ETTm1, ETTm2) \cite{zhou2021informer}.

\begin{table*}[ht]
\caption{Results of the Long-Term Time Series Forecasting task. For all baselines, we adhere to the setting of the iTransformer with an input sequence length of 96. We conducted comparisons among an array of competitive models during various forecast horizons. All the results are averaged from 4 different prediction lengths, that is \{96, 192, 336, 720\}. A lower MSE or MAE indicates a better prediction, we denote the optimal results with boldface  for clarity.}
   \setlength{\tabcolsep}{5pt}
   \renewcommand{\arraystretch}{1}
	\resizebox{\linewidth}{!}{	
\begin{tabular}{c|cc|cc|cc|cc|cc|cc|cc|cc|cc|cc|cc}
 \toprule[1.0pt]
Methods                 & \multicolumn{2}{c|}{PatchMLP} & \multicolumn{2}{c|}{TimeMixer}                     & \multicolumn{2}{c|}{iTransformer} & \multicolumn{2}{c|}{RLinear} & \multicolumn{2}{c|}{PatchTST} & \multicolumn{2}{c|}{Crossformer} & \multicolumn{2}{c|}{TiDE} & \multicolumn{2}{c|}{TimesNet} & \multicolumn{2}{c|}{DLinear} & \multicolumn{2}{c|}{SCINet} & \multicolumn{2}{c}{FEDformer} \\ \midrule[0.5pt]
Metric                  & MSE           & MAE          & \multicolumn{1}{c}{MSE} & \multicolumn{1}{c}{MAE} & MSE             & MAE            & MSE          & MAE          & MSE           & MAE          & MSE             & MAE           & MSE         & MAE        & MSE           & MAE          & MSE          & MAE          & MSE          & MAE         & MSE           & MAE           \\ \midrule[0.5pt]
ETTm1                   & \textbf{0.374}     & \textbf{0.382}    & 0.381                   & 0.395                   & 0.407           & 0.410          & 0.414        & 0.407        & 0.387         & 0.400        & 0.513           & 0.496         & 0.419       & 0.419      & 0.400         & 0.406        & 0.403        & 0.407        & 0.485        & 0.481       & 0.448         & 0.452         \\ \midrule[0.5pt]
ETTm2                   &    \textbf{0.269}           &   \textbf{0.311}          & 0.275                   & 0.323                   & 0.288           & 0.332          & 0.286        & 0.327        & 0.281         & 0.326        & 0.757           & 0.610         & 0.358       & 0.404      & 0.291         & 0.333        & 0.350        & 0.401        & 0.571        & 0.537       & 0.305         & 0.349         \\ \midrule[0.5pt]
ETTh1                   & \textbf{0.438}              &  \textbf{0.429}            & 0.447                   & 0.440                   & 0.454           & 0.447          & 0.446        & 0.434        & 0.469         & 0.454        & 0.529           & 0.552         & 0.541       & 0.507      & 0.458         & 0.450        & 0.456        & 0.452        & 0.747        & 0.647       & 0.440         & 0.460         \\ \midrule[0.5pt]
ETTh2                   & \textbf{0.349}     &  \textbf{0.378}            & 0.364                   & 0.395                   & 0.383           & 0.407          & 0.374        & 0.398        & 0.387         & 0.407        & 0.942           & 0.684         & 0.611       & 0.550      & 0.414         & 0.427        & 0.559        & 0.515        & 0.954        & 0.723       & 0.437         & 0.449         \\ \midrule[0.5pt]
ECL                     &   \textbf{0.171}            &  \textbf{0.265}            & 0.182                   & 0.272                   & 0.178           & 0.270          & 0.219        & 0.298        & 0.216         & 0.304        & 0.244           & 0.334         & 0.251       & 0.344      & 0.192         & 0.295        & 0.212        & 0.300        & 0.268        & 0.365       & 0.214         & 0.327         \\ \midrule[0.5pt]
Traffic                 &    \textbf{0.417}           &  \textbf{0.273}            & 0.484                   & 0.297                   & 0.428           & 0.282          & 0.626        & 0.378        & 0.555         & 0.362        & 0.550           & 0.304         & 0.760       & 0.473      & 0.620         & 0.336        & 0.625        & 0.383        & 0.804        & 0.509       & 0.610         & 0.376         \\ \midrule[0.5pt]
Weather                 & \textbf{0.231}     & \textbf{0.256}    & 0.240                   & 0.271                   & 0.258           & 0.279          & 0.272        & 0.291        & 0.259         & 0.281        & 0.259           & 0.315         & 0.271       & 0.320      & 0.259         & 0.287        & 0.265        & 0.317        & 0.292        & 0.363       & 0.309         & 0.360         \\ \midrule[0.5pt]
Solar-Energy            &   \textbf{0.211}            & \textbf{0.261}    & 0.216                   & 0.280                   & 0.233           & 0.262          & 0.369        & 0.270        & 0.307         & 0.641        & 0.6398          & 0.347         & 0.417       & 0.301      & 0.319         & 0.330        & 0.401        & 0.282        & 0.375        & 0.281       & 0.291         & 0.381         \\ \midrule[0.5pt]
$\mathrm{1^{st} Count}$ & \textbf{8}    & \textbf{8}   & 0                       & 0                       & 0               & 0              & 0            & 0            & 0             & 0            & 0               & 0             & 0           & 0          & 0             & 0            & 0            & 0            & 0            & 0           & 0             & 0        
\\ \bottomrule[1.0pt]    
\end{tabular}
}

\label{tab1}
\end{table*}

\begin{table*}[!ht]
\center
\caption{Ablations of PatchMLP. We replace the components of PatchMLP one by one and explore the performance of different MLPs.  All the results are averaged from 4 different prediction lengths. A check mark \ding{51} and a wrong mark \ding{55} indicate with and without certain components respectively. 
}
\setlength{\tabcolsep}{5pt}
   \renewcommand{\arraystretch}{1}
	\resizebox{0.8\linewidth}{!}{
\begin{tabular}{c|c|c|c|c|cc|cc|cc|cc|cc}
 \toprule[1.0pt]
\multirow{2}{*}{Case} & \multirow{2}{*}{Decompose} & \multirow{2}{*}{MPE}       & \multirow{2}{*}{\begin{tabular}[c]{@{}c@{}}Dot\\ Product\end{tabular}} & \multirow{2}{*}{\begin{tabular}[c]{@{}c@{}}Intra\\ Variable\end{tabular}} & \multicolumn{2}{c|}{ECL} & \multicolumn{2}{c|}{Traffic} & \multicolumn{2}{c|}{Solar-Energy} & \multicolumn{2}{c|}{Weather} & \multicolumn{2}{c}{ETTm1} \\
                      &                            &                            &                                                                        &                                                                           & MSE        & MAE        & MSE          & MAE          & MSE             & MAE            & MSE          & MAE          & MSE         & MAE         \\ \midrule[0.5pt]
\circled{1}                      & \ding{51} & \ding{51} & \ding{51}                                             & \ding{51}                                                &\textbf{0.171} &\textbf{0.265} &\textbf{0.417} &\textbf{0.273} &\textbf{0.211} &\textbf{0.277} &\textbf{0.231} &\textbf{0.256} &\textbf{0.374} &\textbf{0.382}            \\ \midrule[0.5pt]
\circled{2}                      & \ding{51} & \ding{55} & \ding{51}                                             & \ding{51}                                                &0.183 &0.279 &0.431 &0.282 &0.223 &0.288 &0.241 &0.270 &0.383 &0.395             \\ \midrule[0.5pt]
\circled{3}                      & \ding{51} & \ding{51} & \ding{55}                                             & \ding{51}                                                &0.177 &0.271 &0.426 &0.280 &0.219 &0.283 &0.237 &0.262 &0.383 &0.387             \\ \midrule[0.5pt]
\circled{4}                      & \ding{51} & \ding{51} & \ding{51}                                             & \ding{55}                                                &0.179 &0.276 &0.426 &0.284 &0.218 &0.287 &0.242 &0.266 &0.382 &0.393             \\ \midrule[0.5pt]
\circled{5}                      & \ding{55} & \ding{51} & \ding{51}                                             & \ding{51}                                                &0.186 &0.275 &0.429 &0.287 &0.223 &0.288 &0.243 &0.266 &0.384 &0.390             \\ \midrule[0.5pt]
\circled{6}                      & \ding{55} & \ding{55} & \ding{51}                                             & \ding{51}                                                &0.198 &0.284 &0.442 &0.298 &0.233 &0.301 &0.255 &0.274 &0.396 &0.402            \\ \midrule[0.5pt]
\circled{7}                      & \ding{55} & \ding{51} & \ding{55}                                             & \ding{51}                                                &0.193 &0.280 &0.435 &0.295 &0.230 &0.293 &0.252 &0.275 &0.390 &0.396             \\ \midrule[0.5pt]
\circled{8}                      & \ding{55} & \ding{51} & \ding{51}                                             & \ding{55}                                                &0.194 &0.284 &0.439 &0.296 &0.234 &0.301 &0.253 &0.274 &0.395 &0.398             \\ \midrule[0.5pt]
\circled{9}                      & \ding{51} & \ding{51} & \ding{51}                                             & \ding{51}                                                &0.180 &0.275 &0.429 &0.281 &0.219 &0.287 &0.240 &0.265 &0.382 &0.390        \\ 
\bottomrule[1.0pt]         
\end{tabular}       
}                
\label{tab2}
\end{table*}

\begin{figure*}[ht]
	\centering
	\includegraphics[scale=0.35]{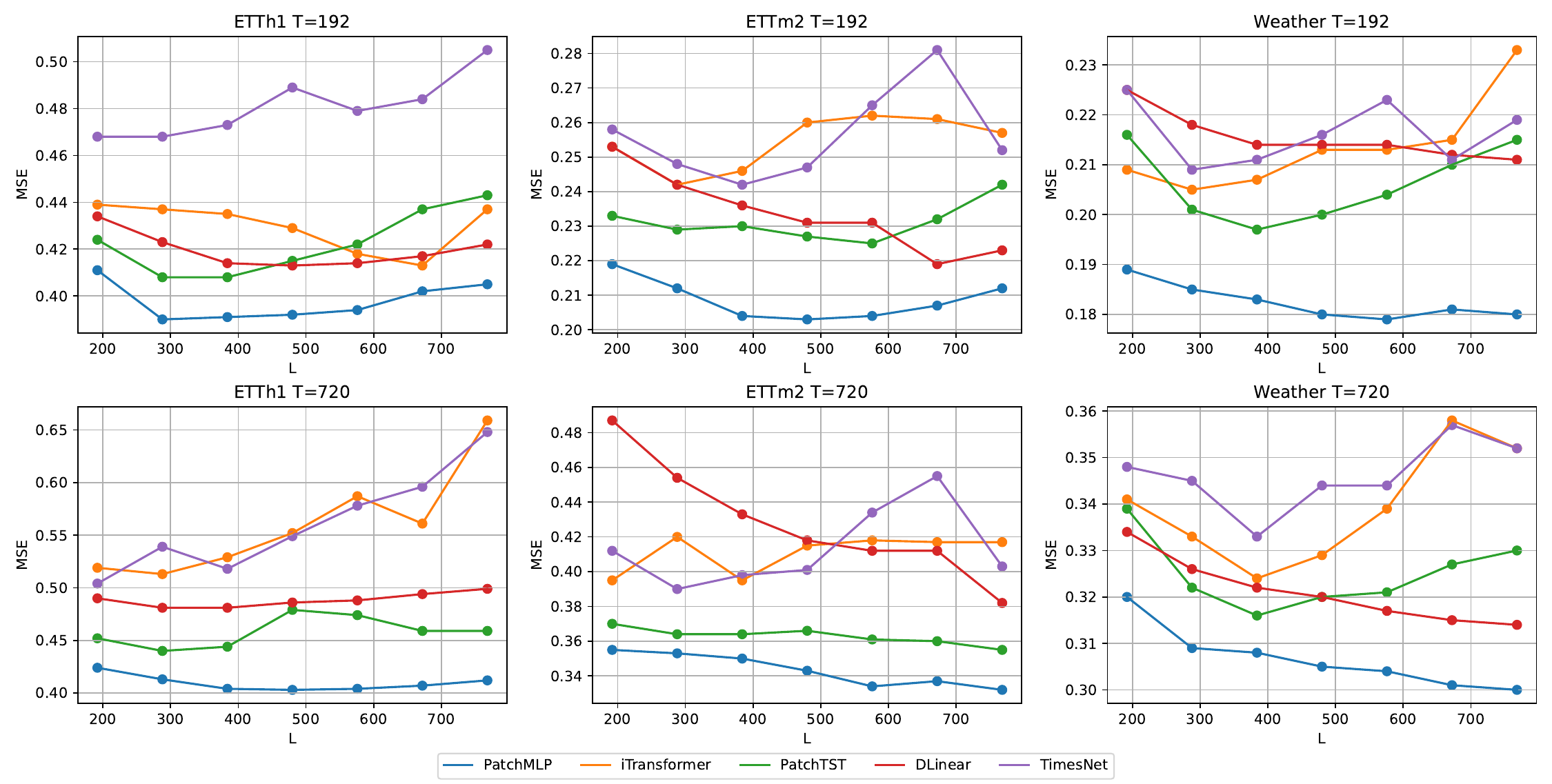}
  \caption{Forecasting performance (MSE) with varying look-back windows on 3 datasets: ETTh1, ETTm2, and Weather. The look-back windows are selected to be $L=\{192, 288, 384, 480, 576, 672, 768\}$, and the prediction horizons are $T = \{192, 720\}$.}
\label{last}
\end{figure*}

\begin{figure*}[!ht]
	\centering
	\includegraphics[scale=0.35]{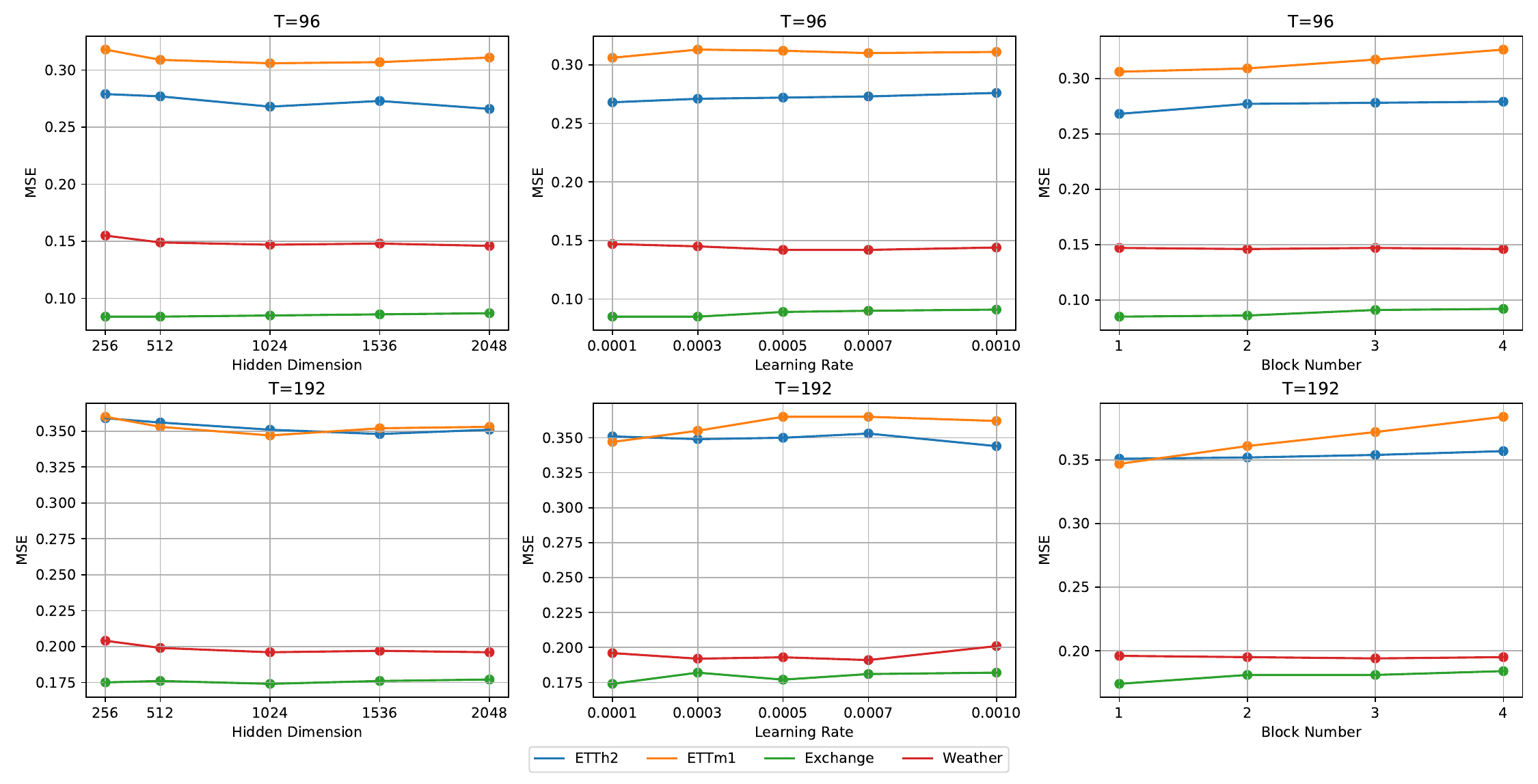}
  \caption{Hyperparameter sensitivity with respect to the learning rate, the number of MLP blocks, and the hidden dimension of Embedding tokens. The results are recorded with the lookback window length $L = 96$ and the forecast window length $T = \{96, 192\}$}
\label{last_ps}
\end{figure*}

\textbf{Baselines and Metrics:} 
We selected nine widely acknowledged SOTA forecasting models for our benchmarking analysis, which comprises the Transformer-based models: iTransformer \cite{liu2023itransformer}, PatchTST \cite{nie2022time}, Crossformer \cite{zhang2022crossformer}, and FEDformer \cite{zhou2022fedformer}; the CNN-based models: TimeNet \cite{wu2022timesnet}, SCINet \cite{liu2022scinet}; and the significant MLP-based  models: Timemixer \cite{wang2024timemixer}, DLinear \cite{zeng2023transformers}, TiDE \cite{das2023long}, and RLinear \cite{li2023revisiting}. To evaluate the performance of these models, we used widely used evaluation metrics: MSE and Mean Absolute Error (MAE).

\subsection{Multivariate Long-Term Forecasting}
Table ~\ref{tab1} presents the results of MTS forecasting, with lower MSE/MAE values indicating greater predictive accuracy. Overall, the model we propose has achieved a surprisingly efficacious outcome, realizing optimal performance across all datasets.
Out of a total of 16 benchmarks, we achieved 100\% state-of-the-art (SOTA) results, surpassing all Transformer architectures. This suggests that the Attention mechanism may not necessarily be the optimal choice for LSTF tasks, as even simple linear models are capable of achieving impressive results. Further analysis indicates that channel independence models, represented by PatchTST, often failed to perform to their potential, thus implying the indispensable role of inter-variable interactions in MTS analysis. Notably, despite explicitly capturing multivariate correlations, the performance of Crossformer did not meet expectations, underscoring that inappropriate utilization of inter-variable correlations can negatively impact model efficacy. Hence, we conclude that skillful leverage of interactions between variables is vital and, in certain scenarios, simple linear models may suffice to deliver outstanding performance compared to the more complex Transformer architectures.

\subsection{Model Analysis}

\subsubsection{Ablation Study}
As shown in the Table ~\ref{tab2}, we investigate the impact of each module of PatchMLP on performance. \circled{9} follows the conventional decomposition approach, which involves decomposing the time series first and then embedding it before separately handling the prediction of the trend and seasonal components. \circled{5} eliminates the decomposition module, directly inputs the embedded original sequence into the model for prediction. In both \circled{2} and \circled{6}, the MPE is removed, and the embedding method is altered to a single linear layer. \circled{3} and \circled{7}, on the other hand, eliminate the dot product in the MLP among variables and replace it with a simple residual connection, while \circled{4} and \circled{8} completely remove the MLP among variables, canceling the interaction between them. We only change the corresponding modules and keep all other settings unchanged, with the input length still set to 96.

It can be observed that conventional decomposition methods do not perform well, which is attributable to the excessively complex temporal relationships inherent in the original time series that cannot be effectively deconstructed through simple decomposition techniques.
In contrast, latent vector decomposition can adeptly circumvent this issue.
Furthermore, the observed performance deterioration upon the removal of the MPE indicates that MPE plays a significant role in learning the various temporal relationships.
Additionally, it is noteworthy that the dot product method outperforms simply because it enhances the interactiveness between variables, while mere addition does not confer this advantage.
Correspondingly, models that lack variable interactiveness experience a predictable decline in performance, which reiterates the critical importance of appropriately leveraging the interrelationships between variables.
\subsubsection{Increasing look-back Window}

In principle, a longer input sequence might enhance model performance, primarily due to the provision of a richer historical context that assists the model in more effectively learning and identifying long-term patterns within time series. 
Generally, a robust LSTF model, equipped with substantial capability to extract temporal relationships, is expected to yield improved outcomes with extended input lengths.
However, longer inputs necessitate the model's heightened ability to capture long-term dependencies, as a deficit in this capacity could readily precipitate a decline in model performance.

As illustrated in Figure ~\ref{last}, there is a gradual augmentation in the performance of all models with increasing input length, but as the input extends to considerable lengths (768), some models begin to exhibit diminished performance, which may also be attributed to the amplified noise accompanying the expanded input.
In contrast, our model and DLinear consistently demonstrate steady improvements in performance as the input lengthens, exemplifying the superiority of linear models in this context.

\subsection{Hyperparameter Sensitivity}

We evaluated the hypersensitivity of the PatchMLP concerning the following parameters: learning rate (lr), the number of blocks (layers) N in the MLP, and the hidden dimension D of the embedding. The results are illustrated in Figure ~\ref{last_ps}. We observed that the performance of PatchMLP is not particularly sensitive to these hyperparameters, as evidenced by the relatively inconspicuous variations in performance. It is advisable that both the number of blocks and the hidden dimension size not be excessively large, which may be attributable to the increase in the total number of parameters that require learning with larger values of these hyperparameters. However, the hidden dimension size should not be excessively small either, to avoid a disproportionate compression of features into an insufficiently capacious latent vector.

\section{Conclusions}
This paper provides an in-depth exploration of existing solutions for Long-Term Time Series Forecasting (LTSF) tasks within the framework of the Transformer architecture and introduces an innovative approach: Patch-based Multi-Layer Perceptron (PatchMLP). Through empirical analysis, we demonstrate the efficacy of the patch mechanism in time series forecasting and address the current limitations of Transformer models by designing a novel, simplified architecture. The PatchMLP utilizes a simple moving average to separate the smooth components from the noisy residuals in time series, and employs a unique channel mixing strategy to enhance the interchange of semantic information across variables. Additionally, we present the Multi-Scale Patch Embedding method to enable the model to more effectively learn the diverse associations among input sequences.
Through extensive experimentation on multiple real-world datasets, PatchMLP shows superior performance compared to existing technologies. This research not only validates the potential of reducing complexity and simplifying network structures to improve performance on LTSF tasks but also emphasizes the critical role of cross-variable interactions in enhancing the accuracy of multivariate time series forecasting .
We anticipate that the success of PatchMLP will not only advance further research in the field of LSTF but also motivate future efforts toward developing models that prioritize efficiency, simplicity, and interpretability. Ultimately, we hope that this research will inspire the creation of more innovative time series forecasting methods focused on addressing specific problems rather than the pursuit of model complexity.

\bibliography{aaai25}

\end{document}